\documentclass[nohyperref]{article}

\usepackage{microtype}
\usepackage{graphicx}
\usepackage{subfigure}
\usepackage{booktabs} 

\usepackage{hyperref}

\usepackage[accepted]{icml2023}

\usepackage{amsmath}
\usepackage{amssymb}
\usepackage{mathtools}
\usepackage{amsthm}

\usepackage{caption}
\usepackage[symbol]{footmisc}

\theoremstyle{plain}

\theoremstyle{definition}

\theoremstyle{remark}

\usepackage[textsize=tiny]{todonotes}
\usepackage{xcolor}

\icmltitlerunning{Oscillation-free Quantization for Low-bit Vision Transformers}

\begin{document}

\twocolumn[
\icmltitle{Oscillation-free Quantization for Low-bit Vision Transformers}



\icmlsetsymbol{equal}{*}
\begin{icmlauthorlist}
\icmlauthor{Shih-Yang Liu}{equal,yyy}
\icmlauthor{Zechun Liu}{equal,comp}
\icmlauthor{Kwang-Ting Cheng}{yyy}
\end{icmlauthorlist}

\icmlaffiliation{yyy}{Hong Kong University of Science and Technology}
\icmlaffiliation{comp}{Reality Labs, Meta Inc}

\icmlcorrespondingauthor{Shih-Yang Liu}{sliuau@connect.ust.hk}
\icmlcorrespondingauthor{Zechun Liu}{zechunliu@fb.com}

\icmlkeywords{Machine Learning, ICML}

\vskip 0.3in
]



\printAffiliationsAndNotice{\icmlEqualContribution} 

\begin{abstract}

Weight oscillation is an undesirable side effect of quantization-aware training, in which quantized weights frequently jump between two quantized levels, resulting in training instability and a sub-optimal final model. We discover that the learnable scaling factor, a widely-used \textit{de facto} setting in quantization aggravates weight oscillation. In this study, we investigate the connection between the learnable scaling factor and quantized weight oscillation and use ViT as a case driver to illustrate the findings and remedies. In addition, we also found that the interdependence between quantized weights in \textit{query} and \textit{key} of a self-attention layer makes ViT vulnerable to oscillation. We, therefore, propose three techniques accordingly: statistical weight quantization ($\rm StatsQ$) to improve quantization robustness compared to the prevalent learnable-scale-based method; confidence-guided annealing ($\rm CGA$) that freezes the weights with \textit{high confidence} and calms the oscillating weights; and \textit{query}-\textit{key} reparameterization ($\rm QKR$) to resolve the query-key intertwined oscillation and mitigate the resulting gradient misestimation. Extensive experiments demonstrate that these proposed techniques successfully abate weight oscillation and consistently achieve substantial accuracy improvement on ImageNet. Specifically, our 2-bit DeiT-T/DeiT-S/Swin-T algorithms outperform the previous state-of-the-art by 9.8\%/7.7\%/4.64\%, respectively. Code and models are
available at: \url{https://github.com/nbasyl/OFQ}.

\end{abstract}

\section{Introduction}
\label{intro}
Deep neural networks have enjoyed tremendous success in numerous applications, ranging from computer vision \cite{he2016deep, dosovitskiy2020image} to natural language processing \cite{vaswani2017attention, devlin2018bert}. However, the prohibitive model size and resource-intensive computation restrict the feasibility of deploying large models on resource-constrained devices. Among many methods that study the compression and acceleration of neural networks \cite{liu2019metapruning, wu2019fbnet, zhou2016dorefa}, quantization-based approaches have stood out due to their high compression ratio and remarkable reduction in throughput time by adopting efficient bitwise operations \cite{zhu2020xor, zhang2018lq}. 

Despite the efficacy of quantization, there is still a non-negligible accuracy gap between quantized models and their full-precision counterparts. The accuracy drop comes from several aspects, including but not limited to the discrete nature of quantization and its limited representational capability~\cite{liu2022nonuniform, zhang2018lq, li2019additive, miyashita2016pot}, difficulty in gradient approximation to the non-differentiable quantization function~\cite{gong2019differentiable, liu2018bi}, and quantization oscillation that hinders optimization~\cite{nagel2022overcoming}. 
The last, relatively under-explored, is the primary focus of this study. 

In this study, we choose the quantized ViT as the driver for investigating the cause of oscillation. Previous work~\cite{nagel2022overcoming} assumes the scaling factors are fixed in quantization, while this is not the case as most of the prevalent quantization methods adopt learnable scaling factors, including \cite{nagel2022overcoming}.
Surprisingly, we found that the learnable scaling factor exacerbates the oscillation, leading to unstable quantization-aware training and often results in sub-optimal models. The learnable scaling factor is updated with noisy gradients and determines quantization thresholds. Thus, the intertwined oscillation between the noisy learnable threshold and the weights near the threshold makes training unstable. 

We visualize the composition of the oscillating weights and observe that they are position-agnostic, meaning that there are no such weights that would persistently stay within the oscillation region during optimization. But the current optimizer fails to prevent new weights from entering the oscillating region while cleaning weights out, so the oscillation cannot subside. Additionally, we find that in a na\"ively quantized self-attention layer, oscillation in quantized \textit{query} will directly impact the gradient estimation for the weights in \textit{key} and vice versa, resulting in \textit{query}-\textit{key} weight co-oscillation. 

To this end, we propose three novel techniques based on our analyses to abate quantization oscillation, namely statistical weight quantization ($\rm StatsQ$) and confidence-guided annealing ($\rm CGA$) to stabilize training and ultimately eliminate the weight oscillation, and \textit{query}-\textit{key} reparameterization ($\rm QKR$) to decouple the negative mutual influence between quantized \textit{query} and \textit{key}.

We demonstrate that the proposed techniques are complementary and capable of working collaboratively to achieve oscillation-free quantized ViTs with consistent improvements over the previous state-of-the-art on ImageNet. Specifically, our quantization method produces 2-bit DeiT-T and DeiT-S that improve the ImageNet top-1 accuracy by 9.88\% and 7.72\% respectively, compared to previous SoTA, and we show for the first time that 3-bit DeiT-T and DeiT-S can achieve comparable accuracy as the full-precision models.

\section{Related Works}
\label{related_works}

Model quantization can be categorized into post-training quantization (PTQ)~\cite{liu2021post, nagel2020up, banner2019post, nagel2019data} and quantization-aware training (QAT)~\cite{zhou2016dorefa, choi2018pact, gong2019differentiable}. In general, PTQ provides a faster quantization pipeline than QAT, as PTQ relies on calibration data without re-training; on the other hand, QAT is more suitable for precision-sensitive scenarios at the cost of lengthier training time. This work focuses on QAT.

Prior literature has tried to ameliorate QAT from different aspects. Some proposed increasing the quantization representation ability by replacing uniform quantization with non-uniform quantization, such as \cite{liu2022nonuniform, li2019additive, zhang2018lq}. Another line of research suggests reducing the gradient misestimation caused by the non-differentiable rounding function by approximating the discrete quantization with a differentiable function \cite{gong2019differentiable, zhou2016dorefa}. Moreover, several works have delved into searching for the optimal scaling factor. For instance, \cite{zhou2016dorefa} incorporates a non-linear function for scaling the weights to restrict their value range. \cite{choi2018pact} introduces a trainable clipping parameter to sort out the suitable quantization scale automatically. Lately, \cite{esser2019learned} introduced a simple, intuitive, yet effective method that resolves around the learnable scaling factor and has been widely adopted as the \textit{de facto} approach for quantization-aware training. 

Recently, \cite{nagel2022overcoming} points out that weight oscillation seriously impacts QAT, mainly rooted in  depth-wise convolution (DW-Conv) and the batch normalization (BN) layers. It uses convolution neural networks (CNNs) for the case study. In our study, we found that in vision transformers (ViTs), despite the absence of DW-Conv and BN, the oscillation still exists, which motivates us to choose ViT as a case driver to investigate quantized weight oscillation. In addition, previous work \cite{nagel2022overcoming} assumed the quantization threshold to be static but neglected the intertwined relation between the learnable scaling factor and weight oscillation. To this end, we delve into studying the entangled relationship between the ViT structure and the oscillation of the quantized weights and understanding the impact of learnable scaling factors on aggravating such oscillation.

\section{Preliminary}
\subsection{Quantization-Aware Training}
In quantization-aware training (QAT), the scaling factor is crucial to striking a good trade-off between the representation value range and the quantization step size, especially in low-bit quantization. Previous works \cite{zhou2016dorefa, choi2018pact, esser2019learned} extensively studied the solution for obtaining a good scaling factor to bridge the accuracy gap between the quantized models and their full precision counterparts. Among these methods, LSQ~\cite{esser2019learned} has become the most prevalent method due to its simplicity and effectiveness in learning the scaling factors. Therefore, in this work, we use \cite{esser2019learned} to study the impact of the learnable scaling factor on the oscillation of QAT.

In LSQ~\cite{esser2019learned}, the quantization function is formulated as:
\begin{equation}
\label{eq:lsq}
    \mathbf{W}_q^{(i)} = \alpha \hat{\mathbf{W}}_q^{(i)} = \alpha \cdot \left\lfloor {\rm Clip} \left( \frac{\mathbf{W}^{(i)}}{\alpha} , Q_n, Q_p \right) \right\rceil
\end{equation}
where $\mathbf{W_q}$ and $\mathbf{W}$ denote the quantized weights and real-valued weights, respectively, and the superscript $i$ denotes the $i^{\rm th}$ entry. $\alpha$ is the learnable scaling factor, and $Q_n$, $Q_p$ represents the quantization range. 

\begin{figure*}[t]
\includegraphics[width=17cm]{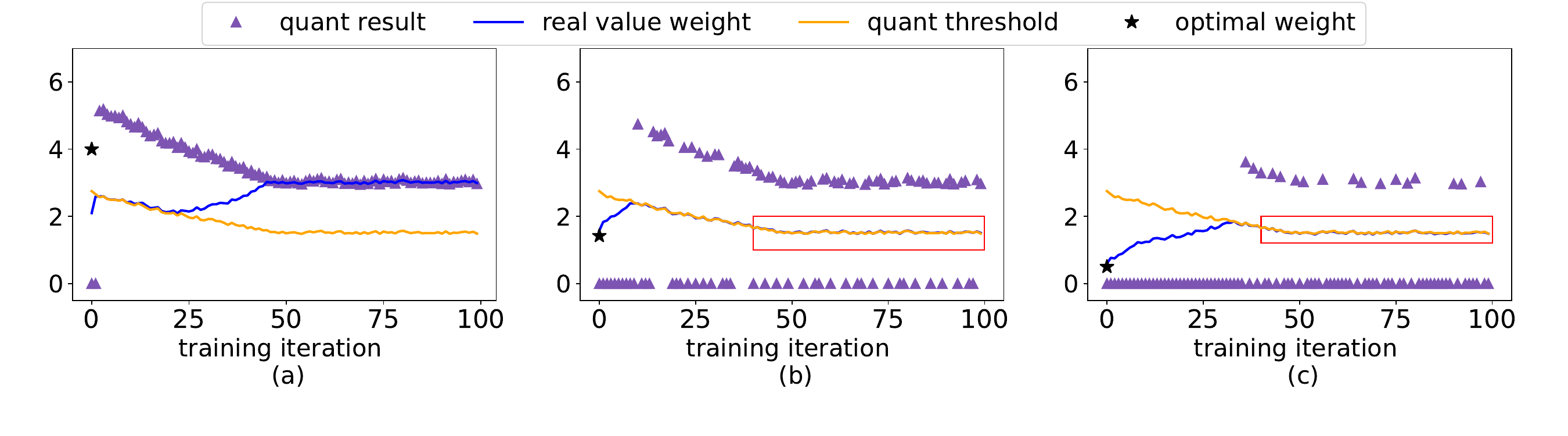}
\caption{The oscillation of each weight (\textit{i.e.}, (a) $\mathbf{W}^{(1)}$ (b) $\mathbf{W}^{(2)}$, and (c) $\mathbf{W}^{(3)}$ ) in a 3D toy-regression example. The red frames show that the weights interweave with the learnable scaling factors and exacerbate the oscillation of both parties. }
\label{fig:toy_example}
\end{figure*}

To approximate the gradient through the non-differentiable round function, straight through estimator (STE) \cite{bengio2013estimating} is adopted: 
\begin{equation}
\label{eq:ste}
    \frac{\partial{\mathcal{L}}}{\partial \mathbf{W}^{(i)}} \overset{STE}{\approx}  \frac{\partial{\mathcal{L}}}{\partial \mathbf{W}_q^{(i)}} \cdot \mathbf{1}_{Q_n\leq \mathbf{W^{(i)}}/\alpha \leq Q_p}
\end{equation}
where $\mathbf{1}$ represents the indicator function that outputs 1 if $\mathbf{W^{(i)}}/\alpha$ is the inside clipping range and 0 otherwise.

\subsection{Quantized ViT Architecture}
\label{sec:qvit_architect}
The main characteristic that differs vision transformer (ViT) from convolutional neural networks (CNNs) is the transformer layer structure, comprising two sub-modules: Multi-head Self-Attention layer (MHSA) and Feed-Forward Network (FFN), which heavily utilize the fully-connected (FC) layers. We adopted row-wise granularity for quantizing weights and activations in FC layers and for quantized activation-activation multiplication in attention layers: 
\begin{align}
\label{eq:qlinear}
    \mathbf{X}_q \cdot \mathbf{Y}_q^{\rm T} = \alpha_{_{\mathbf{X}_q}}\alpha_{_{\mathbf{Y}_q}} \odot ( \hat{\mathbf{X}}_q \otimes \hat{\mathbf{Y}}_q^{\rm T}))
\end{align}
where $\mathbf{X}_q$ and $\mathbf{Y}_q$ denote the quantized tensors, $\otimes$ denotes the integer matrix multiplication and $\odot$ denotes the high-precision scalar-tensor multiplication. Note that, the scaling factor should align with the multiplication direction to make Eq.~\ref{eq:qlinear} established and facilitate low-cost integer matrix multiplication acceleration~\cite{esser2019learned, xiao2022smoothquant}. For example, when input $\mathbf{X}_q \in \mathbb{R}^{N \times D_1}$ and $\mathbf{Y}_q \in \mathbb{R}^{D_2\times D_1}$, we have the scaling factor vector $\alpha_{_{\mathbf{X}_q}} \in \mathbb{R}^{N}$ and $\alpha_{_{\mathbf{Y}_q}} \in \mathbb{R}^{D_2}$. We denote this as quantizing along the last dimension, where the scaling factor is shared along the last dimension. Specifically, in ViTs, weight and activation matrices are all quantized along the last dimension, except for the \textit{value} matrix  $\mathbf{V} \in \mathbb{R}^{N \times D}$ in multiplication with the attention matrix $\mathbf{Attn} \in \mathbb{R}^{N \times N}$, where $\mathbf{V}$ is quantized along the sequence length dimension $N$ and has a scaling factor vector $\alpha_{_{\mathbf{V}_q}} \in \mathbb{R}^{D}$. 

\section{Oscillation in QAT}
\label{problem}
\subsection{Oscillation and Learnable Scaling Factor}
\label{sec:toy_example}

Despite the prevalence of learnable scaling factor in QAT, its negative impact on training stability is rarely studied. 
We illustrate the issues by proposing a toy example with three weights $\mathbf{W} = \{\mathbf{W}^{(1)}, \mathbf{W}^{(2)}, \mathbf{W}^{(3)}\}$ and a scaling factor $\alpha$ determining the quantization threshold for all the three weights, following Eq.~\ref{eq:lsq}. The optimization objective is a 3D regression problem minimizing the $l2$-loss between the target optimal floating-point weight $\mathbf{W_*} \!\in\!\mathbb{R}^{3}$ and quantized weight $\mathbf{W}_q$ in weighting the data vector $\mathbf{X}\!\in\!\mathbb{R}^{3}$ :
\begin{equation}
    \min_{\mathbf{W}} \mathcal{L}(\mathbf{W}) = \mathbb{E}_{\mathbf{X} \sim U} \left[\frac{1}{2}{(\mathbf{X}\mathbf{W_*} - \mathbf{X}\mathbf{W}_q)}^{2} \right] .
\end{equation}
Here $\mathbf{X}$ are sampled from a uniform distribution $U$ on the interval $[0,1)$. We follow Eq.~\ref{eq:ste} to optimize the weight. 

We observe that the weight oscillation caused by quantization's discrete nature can be greatly amplified by the presence of a learnable scaling factor. In the absence of learnable scaling factors, weights farther from their optimal values tend to be updated more towards the target, while weights close to the optima remain stable. However, we found that this is not the case when learnable scaling factors are present. 

As shown in Fig.~\ref{fig:toy_example} (b) and (c), weights ($\mathbf{W}^{(2)}$ and $\mathbf{W}^{(3)}$) initialized close to their target values are influenced by the learnable scaling factor and end up oscillating around the quantization threshold. Note that the scaling factor determines the threshold according to Eq.~\ref{eq:lsq}. The cause of this phenomenon is the existence of an \textit{outlier} weight ($\mathbf{W}^{(1)}$) that is initialized far from its optima ($\mathbf{W}^{(1)}_{*}$). 
The \textit{outlier} weight contributes more to the gradient of the learnable scaling factor, driving it towards an optimal value for this \textit{outlier}, and resulting in the oscillation of the other two weights that are initially set near their optimal values. This observation aligns with the gradient derivation of LSQ~\cite{esser2019learned}: Weights within the range of $[Q_n, Q_p]$ have limited influence on the scaling factor gradient, with values restricted to between -0.5 and 0.5. On the other hand, weights outside of this range contribute significantly more to the gradient of the scaling factor, e.g. scaled weights that are larger than $Q_p$ contribute $Q_p$ to the gradient of the scaling factor. 

More catastrophically, once the weights interweave with the learnable thresholds, as highlighted by the red frame in Fig.~\ref{fig:toy_example}, weight oscillation will introduce noise to the gradient of the learnable scaling factor and cause it to oscillate. The latter, in turn, changes the quantization threshold and aggravates the weight oscillation. This vicious cycle of interaction makes the oscillation hard to subside. 


\subsection{Oscillations in Quantized ViTs}
The aggravation of weight oscillation caused by the learnable scaling factor does not just exist in the toy example. It is a common problem in real-world scenarios. In this section, we choose the vision transformer (ViT) as a case study for quantization oscillations. We analyze weight oscillation in the quantized ViT and examine how the learnable scaling factor exacerbates the oscillation.

\subsubsection{The Effect of Learnable Quantization on Quantized ViTs}
\label{sec:lsq_on_ViT}

\begin{figure}[t]
\begin{center}
\centerline{\includegraphics[width=0.75\columnwidth]{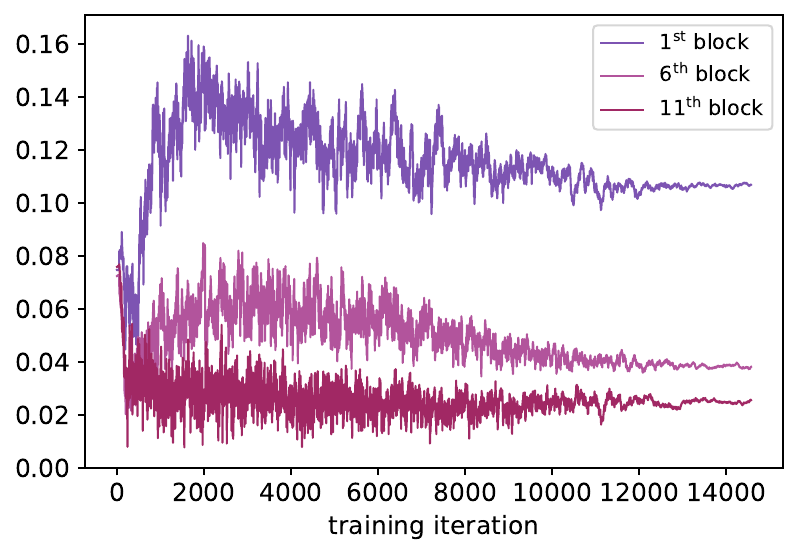}}
\caption{Trajectory of three learnable scaling factors $\alpha$ from the $1^{\rm st}$/$6^{\rm th}$/$11^{\rm th}$ transformer blocks in a 2-bit DeiT-T. The y-axis represents the value of $\alpha$. The fluctuation of learnable scaling factor persists throughout the training.}
\label{fig:lsf_vis}
\end{center}
\end{figure}

\begin{figure}[t]\centering
\minipage{0.24\textwidth}
  \includegraphics[width=\linewidth]{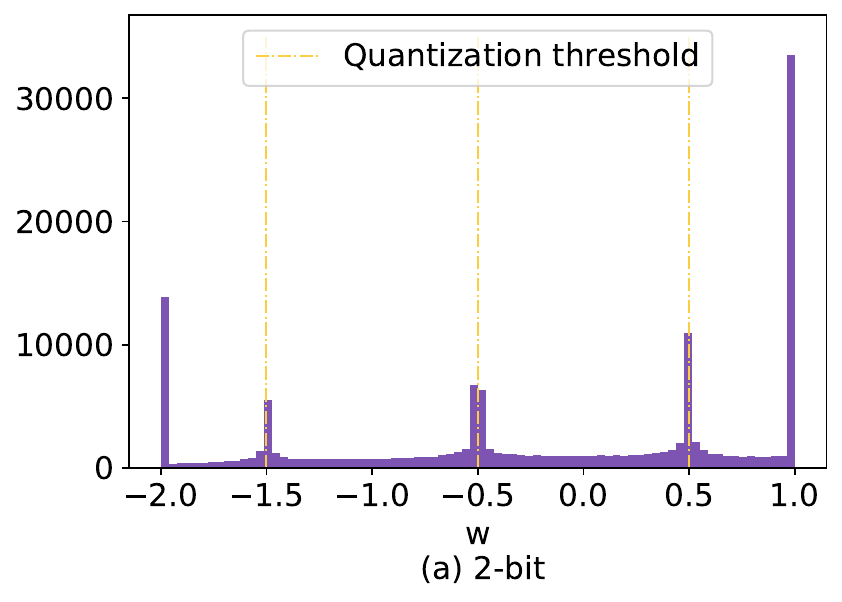}
\endminipage\hfill
\minipage{0.24\textwidth}
  \includegraphics[width=\linewidth]{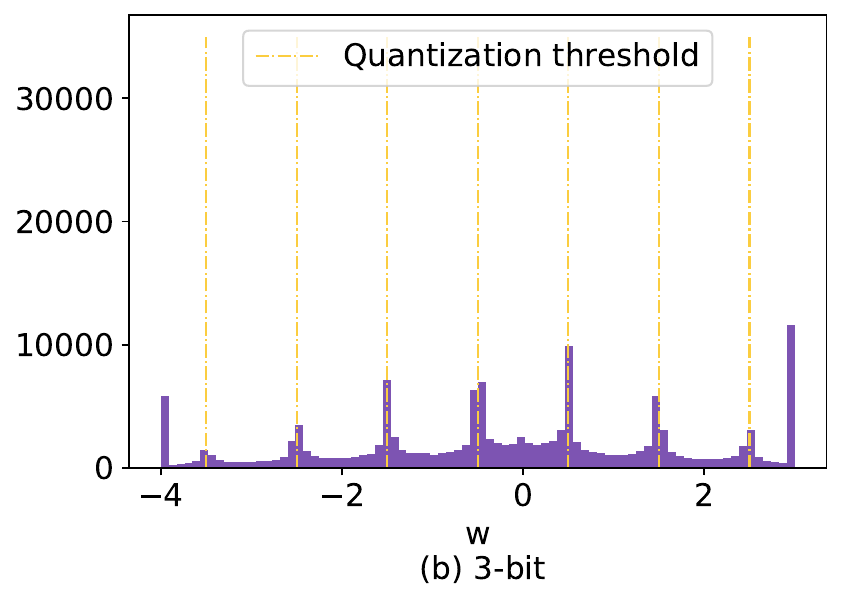}
\endminipage\par
\minipage{0.24\textwidth}
  \includegraphics[width=\linewidth]{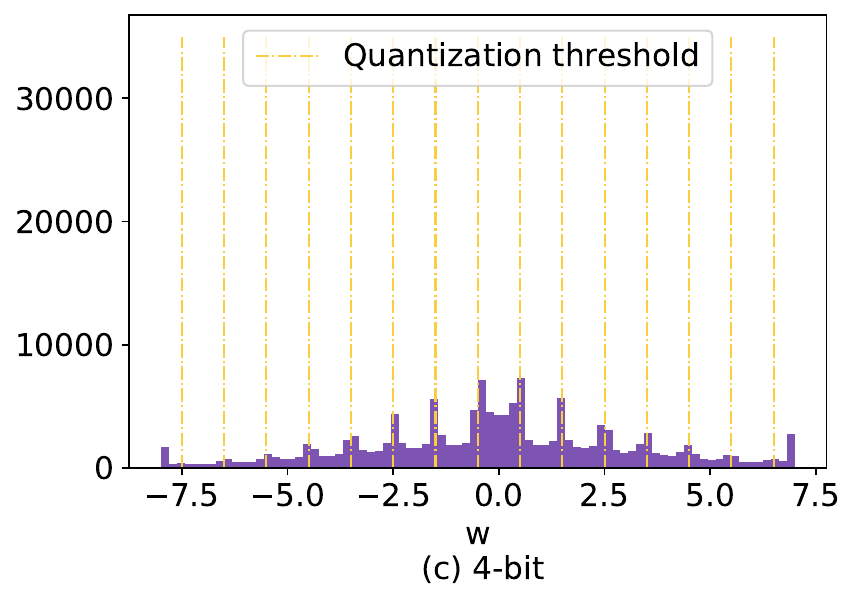}
\endminipage
\caption{Histogram of the weight distribution in the 1st fully connected layer in the feed-forward network (FFN) in the 8th transformer block of the (a) 2-bit, (b) 3-bit, and (c) 4-bit quantized DeiT-T.}\label{fig:weight_vis}
\end{figure}

Throughout the training, we visualize the trajectory of three learnable scaling factors in a quantized DeiT-T~\cite{touvron2021training}. Fig.~\ref{fig:lsf_vis} shows that the scaling factors fluctuated drastically, and the fluctuation persists toward the end of training even when the learning rate is small, and the model is supposed to be converged. From our analysis in Sec.\ref{sec:toy_example}, the oscillation of the learnable scaling factor will contribute to weight oscillation. Fig.~\ref{fig:weight_vis} evidences this by showing clear patterns of a large portion of weights clustered around the quantization threshold. This phenomenon also indicates that there are still non-negligible portions of weights that fail to converge at the end of training and continue to oscillate around the threshold. 

\begin{figure}[t]
\begin{center}
\centerline{\includegraphics[width=0.75\columnwidth]{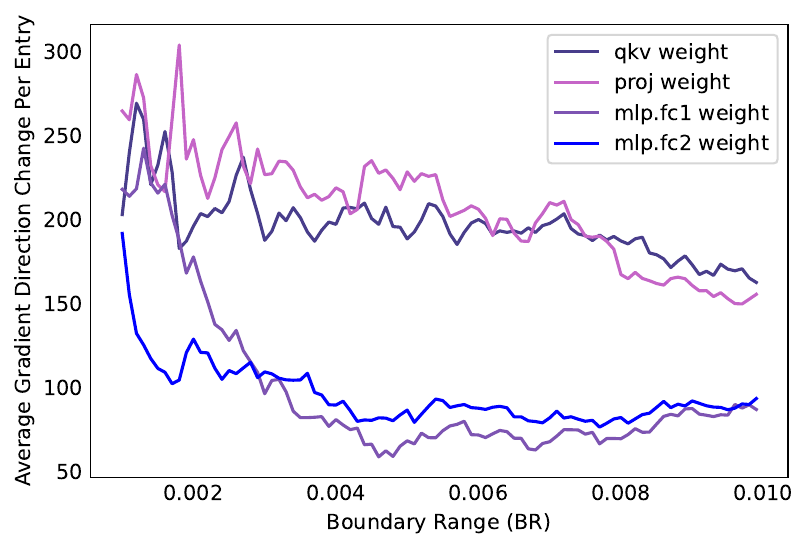}}
\caption{The average number of gradient direction change in 2000 iterations in 2-bit DeiT-T towards the end of training. The frequency of gradient change is anti-correlated with the distance of weights to the quantization thresholds ($\rm BR$).}
\label{fig:gradient_vis}
\end{center}
\end{figure}

We take one step further and inspect the average gradient direction change of weights close to the quantization threshold. We define the term "Boundary Range ($\rm BR_x$)" as a range that includes all weights $\widetilde{\mathbf{W}}_q$ within the distance $x$ to the nearest quantization threshold where $\widetilde{\mathbf{W}}_q = \mathbf{W}/\alpha$ is the rescaled $\mathbf{W}$ before rounding. Fig.~\ref{fig:gradient_vis} clearly shows that the closer the weight entry is to the quantization threshold, the more frequently its gradient direction flips, and such a phenomenon occurs consistently across all modules in ViT. This phenomenon validates that during oscillation, the gradients of weights near the threshold constantly change directions, push the underlying real value weights to cross the threshold, make the quantized weights jump between two quantized values, and prevent the network from well-converging.

\subsubsection{Noise Injection Analysis}
\label{sec:noise_inject}

\begin{table}[!t]
\centering
\caption{Noise injection analysis on quantized DeiT-T. The $1^{\rm st}$, $6^{\rm th}$, and $11^{\rm th}$ row are the accuracies of the converged model before the noise injection. ``Random'' refers to random position noise injection. ``within $\rm BR$'' refers to injecting noise only to weights within the boundary. ``\% of weight'' refers to the fraction of the weights with noise injected. $\mu$ and $\sigma$ denote the mean and variance over ten experiment trials.}
\vspace{-1em}
\setlength{\tabcolsep}{2mm}
\resizebox{0.48\textwidth}{!}{
\begin{tabular}{l|cccr}
\toprule
Method & Val Top-1 & \% of weights \\
\midrule
W2A2 & 54.20\% & -- \\
W2A2 Random ($\mu + \sigma$) & 51.85\%$\pm$0.0873 & 12.15\% \\
W2A2 Random (best) & 52\% & 12.15\% \\
W2A2 (within $\rm BR$) ($\mu + \sigma$) & 54.69\%$\pm$ 0.0803 & 12.15\% \\
W2A2 (within $\rm BR$) (best) & 54.86\% & 12.15\% \\
\midrule
W3A3 & 67.56\% & -- \\
W3A3 Random ($\mu + \sigma$) & 66.72\%$\pm$0.0441 & 7.94\% \\
W3A3 Random (best) & 66.78\% & 7.94\% \\
W3A3 (within $\rm BR$) ($\mu + \sigma$) & 67.67\%$\pm$0.0923 & 7.94\% \\
W3A3 (within $\rm BR$) (best) & 67.83\% & 7.94\% \\
\midrule
W4A4 & 72.58\% & -- \\
W4A4 Random ($\mu + \sigma$) & 72.35\%$\pm$0.0617 & 3.61\% \\
W4A4 Random (best) & 72.422\% & 3.61\% \\
W4A4 (within $\rm BR$) ($\mu + \sigma$) & 72.49\%$\pm$0.0499 & 3.61\% \\
W4A4 (within $\rm BR$) (best) & 72.546\% & 3.61\% \\
\bottomrule
\end{tabular}}
\label{tab:noise_inject_table}
\vskip -0.1in
\end{table}

To further verify our hypothesis that the final quantized network lands at a sub-optima due to weight oscillation within a certain boundary range close to the thresholds, we perform a series of noise injection analyses on quantized DeiT-T. Consider $\rm BR_{0.005}$ as an example. We first inject random noise to weights within $\rm BR_{0.005}$ and train the model for one epoch. The results in Table \ref{tab:noise_inject_table} demonstrated that, surprisingly, randomly injecting noise to weights within $\rm BR_{0.005}$ does not lead to any accuracy degradation on 2-bit and 3-bit models but even improves the model accuracy in the best cases, suggesting that the final model is just a random \textit{snapshot} of the model in oscillation, and there are abundant better solutions near the weight oscillation region. 

Further, to examine if arbitrarily injecting noises to all the weights will have the same effect, we inject random noise to the same proportion of the total weights uniformly, instead of just weights within the $\rm BR$. Predictably, this setting results in a severe accuracy drop, specifically in 2-bit and 3-bit settings. We also observe that there exist fewer weights in the oscillation boundary for models with higher bit-width, and that randomly injecting noise to weights either within $\rm BR_{0.005}$ or at arbitrary positions would result in a slight precision drop on the 4-bit DeiT-T, implying that the oscillation severity is anti-correlated to the bit-width. 
Based on the above discussion, we can confidently conclude that those oscillating weights are the primary factor that keeps the model from converging to good local optima. 

\begin{figure}[t]
\begin{center}
\centerline{\includegraphics[width=0.75\columnwidth]{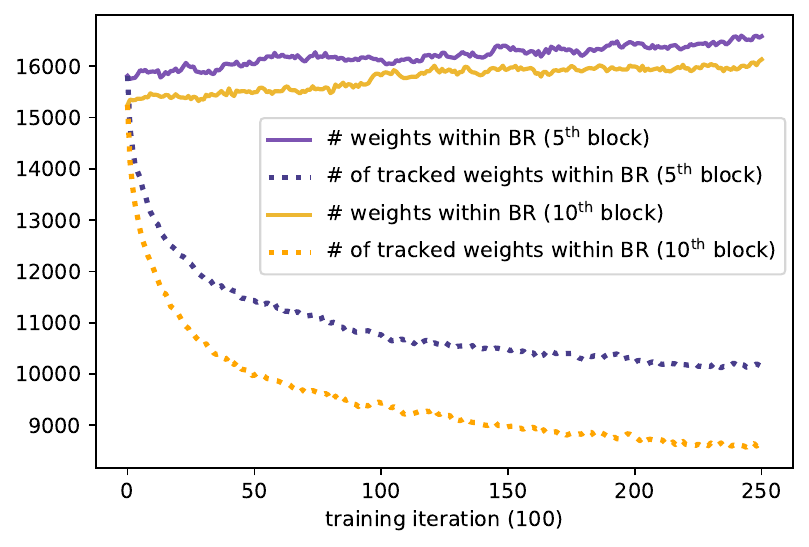}}
\caption{Comparison between the number of weights that is initially within the boundary range $\rm BR_{0.005}$ remaining in the boundary, and the total number of weights within $\rm BR_{0.005}$ of the $5^{\rm th}$/$10^{\rm th}$ transformer block in a 2-bit DeiT-T towards the end of the training.}
\label{fig:oscillation_composition_vis}
\end{center}
\end{figure}

\subsubsection{Composition of the Oscillating Weights}
\label{sec:composition_of_oscillation}
Out of curiosity, we wanted to understand the composition of the oscillating weights, \textit{i.e.,} whether some specific positions in a weight tensor are prone to vibration or oscillation weights are position agnostic. Following the rationale of our previous settings, we tracked weights that were initially within $\rm BR_{0.005}$ in the last few epochs in DeiT-T training. In Fig.~\ref{fig:oscillation_composition_vis}, we notice that the number of tracked weights which continue to stay within $\rm BR_{0.005}$ gradually declines, while the total number of weights within $\rm BR_{0.005}$ is not reduced. This implies that although weights initially in the oscillation region gradually escape during optimization, new weights also enter the boundary region, preventing the network from getting rid of oscillation. This observation also suggests that the oscillating weights are position agnostic, and if we can prevent new weights from entering the boundary range, the oscillation will subside once all the weights are optimized to be outside the oscillation region. This naturally leads to our confidence-guided annealing solution in Sec.~\ref{sec:cga}.

\subsubsection{Bottleneck of Quantized ViTs}
\label{sec:ViT_bottleneck}
Previous literature has observed that quantizing self-attention operations brings the worst accuracy drop compared to other modules\cite{liz2022vit, liy2022q, liz2022qmix}. However, none of those works provides a concrete explanation for that. In the investigation, we found the most quantization-sensitive part of the self-attention lies in the multiplication between quantized \textit{query} ($\mathbf{X_\mathcal{Q}}_q$) and \textit{key} ($\mathbf{X_\mathcal{K}}_q$):
\begin{equation}
\begin{split}
\label{eq:forward_qk_b4}
\!\!\!\mathbf{X\!_\mathcal{Q}}_q\!\!\cdot\!\mathbf{X_\mathcal{K}}^{\rm T}_q\!\!=\!\!F_q(F_q(\mathbf{X}) \!\!\cdot\!\! F_q(\mathbf{W}\!_{\mathcal{Q}}^{\rm T})\!)\!\cdot\!F_q (\!F_q ({\mathbf{W}\!_\mathcal{K}})\!\!\cdot\!\! F_q({\mathbf{X}}^{\rm T})\!)
\end{split}
\end{equation}
Here $F_q$ denotes the quantization function, and $\mathbf{X}$ denotes the input to the attention block. The corresponding back-propagation is formulated as follows:
\begin{equation}
\label{eq:backward_qk_b4}
    \!\!\!\!\!\frac{\partial \mathcal{L}}{\partial \mathbf{W}_\mathcal{K}} \!\!\overset{STE}{\approx}\!\! \frac{\partial \mathcal{L}}{\partial \mathbf{X}_{out}} \!\cdot\! F_q(F_q(\mathbf{X}) \!\cdot\! F_q(\mathbf{W}_{\mathcal{Q}}^{\rm T})) \!\cdot\!  F_q(\mathbf{X}^{\rm T})
\end{equation}
where $\mathbf{X}_{out} = \mathbf{X\!_\mathcal{Q}}_q \cdot \mathbf{X\!_\mathcal{K}}_q$. For simplicity, the terms $\mathbf{1}_{Q_n\leq F_q ({\mathbf{W}_\mathcal{K}}) \cdot F_q({\mathbf{X}}^{\rm T})/\alpha \leq Q_p}$ and $\mathbf{1}_{Q_n\leq \mathbf{W}_\mathcal{K}/\alpha \leq Q_p}$ are omitted here. The gradient \textit{w.r.t.} $\mathbf{W}_\mathcal{K}$ depends on $F_q(F_q(\mathbf{X}) \!\cdot\! F_q(\mathbf{W}_{\mathcal{Q}}^{\rm T}))$ and $F_q(\mathbf{X}^{\rm T})$. Thus, $\mathbf{W}_\mathcal{K}$ optimization would be affected by the inaccurate estimation from oscillation in $F_q(\mathbf{W}_\mathcal{Q})$, and the resulting fluctuation in $F_q(\mathbf{W}_\mathcal{K})$ will, in turn, aggravate $F_q(\mathbf{W}_\mathcal{Q})$ oscillation. To decouple this oscillation aggravation loop, we propose re-parameterizing quantized \textit{query-key} multiplication in Sec. \ref{sec:qkr}.

\section{Conquering Oscillation in Quantized ViT}
\label{methods}
We propose three novel techniques to abate quantization oscillations based on the above observations.    

\subsection{Statistical Weight Quantization}
To eliminate the disproportionate influence from \textit{outlier} weights to learnable scaling factor and reduce the oscillation interweave in between weights and scales, we propose statistic-based weight quantization $\rm StatsQ$:
\begin{equation}
\begin{split}
\label{eq:stats_q}
{\mathbf{W}}_q & = \alpha_s \cdot ((\left\lfloor {\rm Clip} ( \frac{\mathbf{W}}{\alpha_s},-1,1) \cdot n - \! 0.5  \right\rceil \! + \! 0.5) \cdot \! \frac{1}{n}) \\
& {\rm where} \ \ \alpha_s = 2 \cdot \frac{||\mathbf{W}||_{\rm l1}}{|\mathbf{W}|}, \ n  = 2^{b-1}
\end{split}
\end{equation}
Inspired by~\cite{liu2022nonuniform}, $\alpha_s$ is calculated based on the maximum information entropy theory that when the quantized weights are evenly distributed in quantized levels, the highest entropy is preserved. Furthermore, the ${\rm Clip}$ function disregards the \textit{outliers} in scaled weights, and the statistical calculation of scaling factor $\alpha_s$ equally counts the contribution from each weight in the update.

\begin{figure}[t]
\begin{center}
\centerline{\includegraphics[width=0.75\columnwidth]{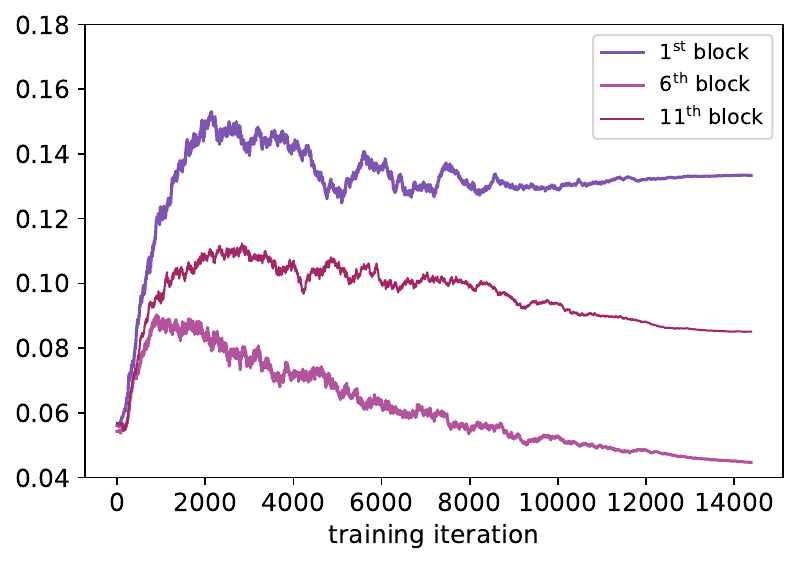}}
\caption{Trajectory of three statistical scaling factors $\alpha_s$ from the $1^{\rm st}$/$6^{\rm th}$/$11^{\rm th}$ transformer blocks in a 2-bit DeiT-T throughout training. The y-axis represents the value of $\alpha_s$.}
\label{fig:ssf_vis}
\end{center}
\end{figure}

We further visualize the statistical scaling factor $\alpha_s$ throughout training in Fig.~\ref{fig:ssf_vis}. In comparison to Fig.~\ref{fig:lsf_vis}, we clearly see that $\alpha_s$ presents a much smoother progression and is almost steady toward the end of the training, demonstrating fewer fluctuations compared to the learnable scaling factors, and improves the overall QAT stability.

\subsection{Confidence-Guided Annealing}
\label{sec:cga}
Calming oscillations requires weights outside the volatility region to stop entering while optimizing out the weights that oscillate initially. Therefore, we propose confidence-guided annealing ($\rm CGA$), which fine-tunes the final model by freezing weights outside the oscillation region and only updating the weights inside. The weight freezing happens every iteration, and once the weights escape the oscillation region, they are frozen, \textit{i.e.}, not updated anymore. $\rm CGA$ is employed for $n$ fine-tuning epochs following the completion of the regular training phase. The detailed implementation of $\rm CGA$ is summarized in Alg.~\ref{alg:cga}. 

The intuition of $\rm CGA$ also aligns with the weight confidence proposition in \cite{helwegen2019latent, liu2021adam}: the confidence level of a quantized weight is proportional to the distance of its real-valued latent weights to the closest threshold. Weights far from the quantization threshold are considered as possessing ``\textit{high confidence}'', while those oscillating weights close to the threshold are in ``\textit{low confidence}''. In this sense, $\rm CGA$ freezes the ``\textit{high confident}'' weights so that the ``\textit{less confident}'' weights can be optimized with those fixed anchors as a prerequisite and gradually get out of the oscillation boundary. In comparison, the $\rm iterative$ $\rm freezing$ method proposed by \cite{nagel2022overcoming} freezes oscillating weights instead of the non-oscillating weights. We have tried $\rm iterative$ $\rm freezing$ on quantized DeiT-T and do not see any significant accuracy improvement. Thus, we argue that ``\textit{high confident}'' weights should be frozen before the ``\textit{low confident}'' weights, not the other way around, to facilitate better update direction estimation for the oscillating weights.

\begin{figure}[t]
\hspace*{-0.5cm}
\minipage{0.5\textwidth}
  \includegraphics[width=\linewidth]{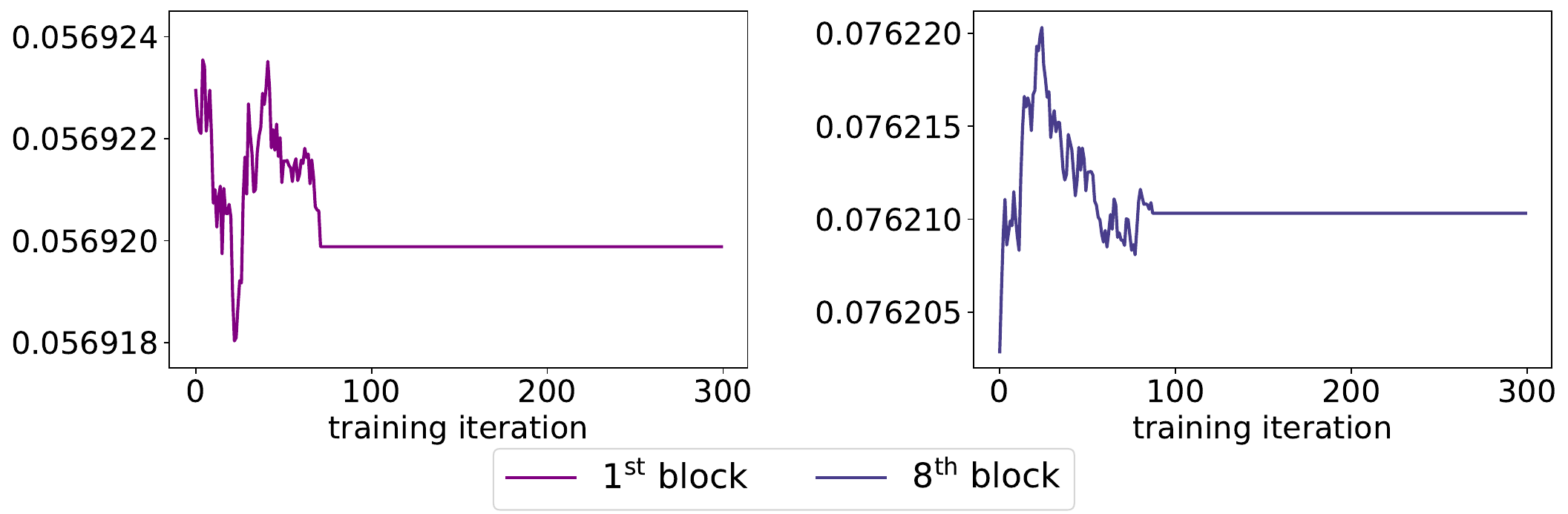}
\endminipage

\caption{ Trajectory of statistical scaling
factors from the $1^{\rm st}$/$8^{\rm th}$ transformer block in a 2-bit DeiT-T throughout the Confidence-Guided Annealing training period. The y-axis represents the value of $\alpha_s$. After a certain number of iterations, statistical scaling
factors stop fluctuating, suggesting all weights have escaped the boundary range and are frozen.}\label{fig:CGA_vis}
\end{figure}

To further validate this point, we visualize the trajectories of two scaling factors $\alpha_{\rm s}$ from the $1^{\rm st}$/$8^{\rm th}$ transformer block in a 2-bit DeiT-T throughout the confidence-guided annealing period. Fig.~\ref{fig:CGA_vis} depicts that the fluctuation of the scaling factor $\alpha_{\rm s}$ ceases after a certain number of iterations. Since $\alpha_{\rm s}$ is calculated with the statistics of the entire weight tensor, the cooling down of $\alpha_{\rm s}$ reflects that all weights have successfully escaped the boundary range and are frozen, resulting in an \textit{oscillation-free} model. Moreover, the top-1 accuracy of a 2-bit quantized DeiT-T is improved from 62.17\% to 64.33\% on ImageNet 1K by applying the $\rm CGA$, which demonstrates the effectiveness of the proposed method. 

\begin{algorithm}
\small
\begin{algorithmic}
   \STATE {\bfseries Given:} Weight Matrix $\mathbf{W}$, Boundary Range ${\rm BR}_{\rm x}$, Annealing Iterations $n$
   \FOR{$j=1$ {\bfseries to} $n$}
   \STATE Calculate gradient $\mathbf{G}_{j} = \frac{\partial \mathcal{L}_{j}}{\partial \mathbf{W}}$ 
   \STATE $\mathbf{G}'_{j} = \mathbf{G}_{j} \cdot \mathbf{1}_{\widetilde{\mathbf{W}}_q \in {\rm BR}_{\rm x}}$
   \STATE \ \ \ \ where $\widetilde{\mathbf{W}}_q = {\rm Clip} ( \frac{\mathbf{W}}{\alpha_s},-1,1) \cdot n - \! 0.5$
   \STATE Update $\mathbf{W}$ with $\mathbf{G}'_{j}$ 
   \ENDFOR
\end{algorithmic}
\caption{Confidence-Guided Annealing}
\label{alg:cga}
\end{algorithm}

\subsection{\textit{\textbf{Q}uery}-\textit{\textbf{K}ey} Reparametrization}
\label{sec:qkr}
Finally, we propose \textit{query}-\textit{key} reparameterization ($\rm QKR$) to decouple the negative mutual-influence between quantized \textit{query} and \textit{key} weights ($\mathbf{W_\mathcal{Q}}_q$, $\mathbf{W_\mathcal{K}}_q$) oscillation. Simple but non-trivially, we reorder the multiplication of $\mathbf{X}_\mathcal{Q}$ and $\mathbf{X}_\mathcal{K}$ in a self-attention module as follows:
\begin{align}
\mathbf{X}_{out} = \mathbf{X}_\mathcal{Q} \cdot \mathbf{X}_\mathcal{K}^{\rm T} = \mathbf{X} \cdot (\mathbf{W}_\mathcal{Q}^{\rm T} \cdot \mathbf{W}_\mathcal{K}) \cdot \mathbf{X}^{\rm T}
\end{align}
Then with the multiplication reordering, the quantization can be formulated as follows:
 \begin{align}
    \label{eq:forward_qk_reparam}
    \mathbf{X}_{out} =  F_q(\mathbf{X})\cdot F_q(F_{q}(\mathbf{W}_{\mathcal{Q}}^{\rm T}\cdot{\mathbf{W}_\mathcal{K}})\cdot F_q({\mathbf{X}}^{\rm T}))
 \end{align}
Here $F_q$ denotes the quantization function. 
The corresponding back-propagation is formulated as:
\begin{equation}
\label{eq:backward_qk_reparam}
    \frac{\partial \mathcal{L}}{\partial \mathbf{W}_\mathcal{K}} \!\!\overset{STE}{\approx}\!\! \frac{\partial \mathcal{L}}{\partial \mathbf{X}_{out}} \cdot F_q(\mathbf{X}) \cdot 
 \mathbf{W}_{\mathcal{Q}}^{\rm T} \cdot F_q(\mathbf{X}^{\rm T})
\end{equation}
Similarly, the terms $\mathbf{1}_{Q_n\leq F_{q}(\mathbf{W}_{\mathcal{Q}}^{\rm T} \cdot {\mathbf{W} _\mathcal{K}}) \cdot F_q({\mathbf{X}} ^{\rm T}) / \alpha \leq Q_p}$ and $\mathbf{1}_{Q_n\leq \mathbf{W}_{\mathcal{Q}}^{\rm T}\!\cdot\!{\mathbf{W}\!_\mathcal{K}}  / \alpha \leq Q_p}$ are omitted here for simplicity. Compared to Eq.~\ref{eq:backward_qk_b4}, in this new gradient derivation \textit{w.r.t.} $\mathbf{W}_\mathcal{K}$, the quantized query weight $F_q(\mathbf{W}_{\mathcal{Q}})$ no longer exists such that the weight co-oscillation in the quantized \textit{query}-\textit{key} multiplication layers is decoupled. Besides, the proposed \textit{query}-\textit{key} reparameterization reduces the number of quantization operations from originally six in Eq.~\ref{eq:forward_qk_b4} to four in Eq.~\ref{eq:forward_qk_reparam}, which lessens the information loss resulting from discretization in the forward pass and estimates more accurate gradients in the backward pass. 

\section{Experiments}
\label{experiments}
In this section, we evaluate our proposed methods on the DeiT-tiny, DeiT-small~\cite{touvron2021training} and Swin-tiny~\cite{liu2021swin} architectures on ILSVRC12 ImageNet classification dataset \cite{imagenet2012}.

\subsection{Implementation Details}
We utilize the official implementation of DeiT\footnote[2]{\url{https://github.com/facebookresearch/deit}} and Swin\footnote[4]{\url{https://pytorch.org/vision/0.13/models/swin_transformer.html}} Our quantized models are trained for 300 epochs with knowledge distillation using the corresponding full-precision models as the teacher models and as initialization. For quantized DeiT-T, 2-bit DeiT-S and 2-bit Swin-T, the training setting follows that of DeiT \cite{touvron2021training} while without mixup/cutmix~\cite{zhang2017mixup, yun2019cutmix} data augmentation. For 3-bit/4-bit quantized DeiT-S and Swin-T, we follow the training recipe in \cite{liy2022q, liz2022qmix}. The number of annealing epochs is set to 25 for fine-tuning the optimized model with $\rm CGA$. We apply 8-bit quantization for the first (patch embedding) layer and the last (classification and distillation) layers following \cite{esser2019learned, liy2022q, liz2022qmix}.

\subsection{Main Results}

\begin{table}[!ht]
\centering
\caption[Caption for LOF]{Comparison of the proposed ${\rm OFQ}$ to previous ViT quantization methods on the ImageNet-1K dataset.}
\setlength{\tabcolsep}{2mm}
\resizebox{0.95\columnwidth}{!}{
\begin{minipage}{1.1\columnwidth}
\begin{tabular}{c|ccc}
\toprule
Network & Method              & Bit-width & Top-1 \% \\ \midrule
    & Full Precision      & W32A32           &     72.02     \\ \cmidrule{2-4} 
                            & $\rm LSQ$~\cite{esser2019learned}    & W2A2             &     54.45     \\
                            & ${\rm QViT}^{\mathbf{*}}$~\cite{liy2022q}      & W2A2             &     50.37     \\
                            & $\mathbf{OFQ}$ \textbf{(Ours)}                 & \textbf{W2A2}             &    \textbf{64.33}    \\ \cmidrule{2-4} 
                            & $\rm LSQ$~\cite{esser2019learned}         & W3A3             &      68.09    \\
                            & Mix-${\rm QViT}$~\cite{liz2022qmix}           & $\sim$W3A3             &      69.62    \\
                {DeiT-T}    & ${\rm QViT}^{\mathbf{*}}$~\cite{liy2022q}      & W3A3             &    67.12      \\
                            & $\mathbf{OFQ}$ \textbf{(Ours)}                  & \textbf{W3A3}             &      \textbf{72.72}    \\ \cmidrule{2-4} 
                            & $\rm LSQ$~\cite{esser2019learned}               & W4A4             &     72.46     \\
                            & Mix-${\rm QViT}$~\cite{liz2022qmix}            & $\sim$W4A4             &     72.79     \\
                            &  ${\rm QViT}^{\mathbf{*}}$~\cite{liy2022q}     & W4A4             &      71.63    \\
                            & $\mathbf{OFQ}$ \textbf{(Ours)}                 & \textbf{W4A4}             &    \textbf{75.46}  \\ \midrule
        & Full Precision      & W32A32           &     79.9     \\ \cmidrule{2-4} 
                            & $\rm LSQ$~\cite{esser2019learned}                 & W2A2       &      68    \\
                            & ${\rm QViT}^{\mathbf{*}}$~\cite{liy2022q}      & W2A2             &     68.67     \\
                            & $\mathbf{OFQ}$ \textbf{(Ours)}                  &\textbf{W2A2}             &    \textbf{75.72} \\ \cmidrule{2-4} 
                            & $\rm LSQ$~\cite{esser2019learned}                 & W3A3             &     77.76     \\
                            & Mix-${\rm QViT}$~\cite{liz2022qmix}           & $\sim$W3A3             &     78.08     \\
         {DeiT-S}           & ${\rm QViT}^{\mathbf{*}}$~\cite{liy2022q}      & W3A3             &     78.45     \\
                            & $\mathbf{OFQ}$ \textbf{(Ours)}                 & \textbf{W3A3}             &     \textbf{79.57}      \\ \cmidrule{2-4}
                            & $\rm LSQ$~\cite{esser2019learned}                 & W4A4             &     79.66     \\
                            & Mix-${\rm QViT}$~\cite{liz2022qmix}           & $\sim$W4A4             &     80.11     \\
                            & ${\rm QViT}^{\mathbf{*}}$~\cite{liy2022q}      & W4A4             &     80.33     \\
                            & $\mathbf{OFQ}$ \textbf{(Ours)}                  & \textbf{W4A4}             &     \textbf{81.10}    
      \\ \midrule
        & Full Precision      & W32A32           &     81.2     \\ \cmidrule{2-4} 
                            & $\rm LSQ$~\cite{esser2019learned}                 & W2A2       &      70.40    \\
                            & ${\rm QViT}^{\mathbf{*}}$~\cite{liy2022q}      & W2A2             &     73.88     \\
                            & $\mathbf{OFQ}$ \textbf{(Ours)}                  &\textbf{W2A2}             &    \textbf{78.52} \\ \cmidrule{2-4} 
                            & $\rm LSQ$~\cite{esser2019learned}                 & W3A3             &     78.96     \\
                            & Mix-${\rm QViT}$~\cite{liz2022qmix}           & $\sim$W3A3             &     79.45     \\
         {Swin-T}           & ${\rm QViT}^{\mathbf{*}}$~\cite{liy2022q}      & W3A3             &     80.06     \\
                            & $\mathbf{OFQ}$ \textbf{(Ours)}                 & \textbf{W3A3}             &     \textbf{81.09}      \\ \cmidrule{2-4}
                            & $\rm LSQ$~\cite{esser2019learned}                 & W4A4             &     80.47     \\
                            & Mix-${\rm QViT}$~\cite{liz2022qmix}           & $\sim$W4A4             &     80.59     \\
                            & ${\rm QViT}^{\mathbf{*}}$~\cite{liy2022q}      & W4A4             &     81.29     \\
                            & $\mathbf{OFQ}$ \textbf{(Ours)}                  & \textbf{W4A4}             &     \textbf{81.88}    
      \\ \bottomrule 
\end{tabular}
\end{minipage}
}
\label{tab:main_table}
\end{table}

We name the combination of our methods \textit{\textbf{O}scillation-\textbf{F}ree \textbf{Q}uantization} ($\rm OFQ$) and present the overall performance on DeiT-T, DeiT-S and Swin-T and compare the result with baseline $\rm LSQ$~\cite{esser2019learned}, Mix-${\rm QViT}$~\cite{liz2022qmix} which proposed a mix-precision quantized ViT, and ${\rm QViT}$\footnote[3]{We have confirmed with the authors of \cite{liy2022q} that their implementation could not establish Eq. \ref{eq:qlinear} due to the reason discussed in Sec. \ref{sec:qvit_architect}. Therefore, we fixed their implementation, reran their experiments following their settings, and reported the results as ${\rm QViT}^{\mathbf{*}}$ in Table \ref{tab:main_table}.}~\cite{liy2022q} in 2/3/4-bit quantization.

\begin{table}[t]
\centering
\caption{Ablation study on the individual effectiveness of the proposed statistical weight quantization ($\rm StatsQ$), confidence-guided annealing ($\rm CGA$), and QK reparameterization ($\rm QKR$) on a quantized DeiT-S.}
\setlength{\tabcolsep}{3mm}
\resizebox{0.95\columnwidth}{!}{%
\begin{tabular}{c|ccc}
\toprule
Network & Method              & Bit-width & Top-1 \% \\ \midrule
                            & Baseline ($\rm LSQ$)                 & W2A2       &      68    \\
                            & ${\rm StatsQ}$           & W2A2             &      74.3    \\
                            & ${\rm StatsQ}$ + ${\rm QKR}$ & W2A2             &    75.00      \\
                            & ${\rm StatsQ}$ + ${\rm CGA}$      & W2A2             &    75.07      \\
                            & ${\rm StatsQ}$ + ${\rm QKR}$ + ${\rm CGA}$        & W2A2             &      \textbf{75.72}    \\ \cmidrule{2-4} 
                            & Baseline ($\rm LSQ$)                  & W3A3             &     77.76     \\
                            & ${\rm StatsQ}$           & W3A3             &     78.56     \\
    {DeiT-S}                & ${\rm StatsQ}$ + ${\rm QKR}$ & W3A3             &     79.15     \\
                            & ${\rm StatsQ}$ + ${\rm CGA}$       & W3A3             &     79     \\
                            & ${\rm StatsQ}$ + ${\rm QKR}$ + ${\rm CGA}$        & W3A3             &     \textbf{79.57}    \\ 
                            \cmidrule{2-4} 
                            & Baseline ($\rm LSQ$)                  & W4A4             &     79.66     \\
                            & ${\rm StatsQ}$           & W4A4             &     80.68     \\
                            & ${\rm StatsQ}$ + ${\rm QKR}$ & W4A4             &     81.00     \\
                            & ${\rm StatsQ}$ + ${\rm CGA}$       & W4A4             &      80.73    \\
                            & ${\rm StatsQ}$ + ${\rm QKR}$ + ${\rm CGA}$        & W4A4             &    \textbf{81.10}   \\ \bottomrule
\end{tabular}%
}
\label{tab:ablation1}
 \vspace{-5pt}
\end{table}

\begin{table}[!h]
\centering
\caption{Ablation study on the selection of boundary ranges (${\rm BR}_{\rm i}$) when applying ${\rm CGA}$. The experiments are conducted using a 2-bit quantized DeiT-S.}
\setlength{\tabcolsep}{3mm}
\resizebox{0.9\columnwidth}{!}{%
\begin{tabular}{c|ccc}
\toprule
Network & Method              & Bit-width & Top-1 \% \\ \midrule
                            & without ${\rm CGA}$     & W2A2      &     75    \\
                            & ${\rm CGA}$  (${\rm BR}_{0.003}$)    & W2A2      &        75.57  \\
   {DeiT-S}                 & ${\rm CGA}$  (${\rm BR}_{0.005}$)     & W2A2      &       \textbf{75.72}   \\
                            & ${\rm CGA}$   (${\rm BR}_{0.007}$)     & W2A2     &      75.63    \\
                            & ${\rm CGA}$   (${\rm BR}_{0.01}$)    & W2A2    &    75.67   \\ \bottomrule
                            \end{tabular}%
}
\label{tab:ablation2}
\vspace{-5pt}
\end{table}

From Table \ref{tab:main_table}, we can first observe that the performance degradation is more severe in lower bit quantization, that is, from real value to 2-bit quantization, the accuracy decreases by 17.57\%, 11.9\% and 10.8\%, respectively on DeiT-T, DeiT-S and Swin-T using the baseline $\rm LSQ$ quantization. In addition, quantized DeiT-T suffers a greater performance drop compared to quantized DeiT-S and Swin-T. The above two observations are consistent with \citep{nagel2022overcoming}'s finding that models of smaller sizes and with lower bit-width exhibit a more severe accuracy drop due to oscillation. In comparison, the proposed $\rm OFQ$ eliminates the oscillation in the final model and thus substantially improves previous SoTAs and narrows the accuracy gap between the quantized model and their full-precision counterparts. Specifically, the 2-bit DeiT-S/Swin-T quantized with $\rm OFQ$ achieves 7.05\%/4.64\% higher accuracy than the previous state-of-the-art ${\rm QViT}$ \cite{liy2022q}, reducing the gap with real-valued models to only 4.18\%/2.68\%. Similarly, the 3-bit OFQ DeiT-T/DeiT-S/Swin-T significantly outperform the 3-bit models in ${\rm QViT}$ \cite{liy2022q} and Mix-${\rm QViT}$~\cite{liz2022qmix}.

In the 4-bit setting, we observe that LSQ can achieve comparable or even higher accuracy than full-precision counterparts, and most previous methods have already surpassed the full-precision models, which implies that the negative effect of oscillation in the 4-bit setting is less detrimental due to higher resolution of the model. Although the room for improvement becomes smaller, $\rm OFQ$ still consistently outperforms all previous works and the accuracy of full-precision models by 3.24\%, 1.2\%, and 0.68\% on DeiT-T, DeiT-S, and Swin-T respectively. 

\subsection{Ablation Study}
\label{sec:ablation}
\begin{figure}[t]
\begin{center}
\minipage{0.5\textwidth}
  \includegraphics[width=\linewidth]{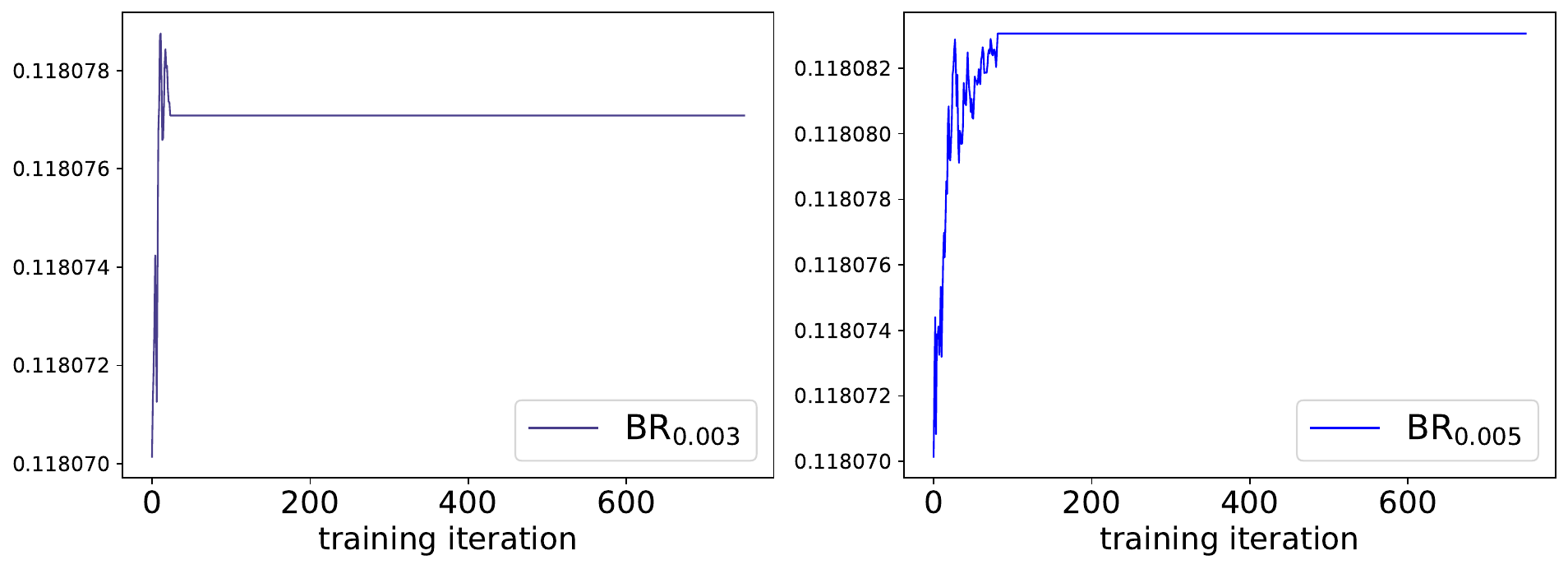}
\endminipage \hfill
\minipage{0.5\textwidth}
  \includegraphics[width=\linewidth]{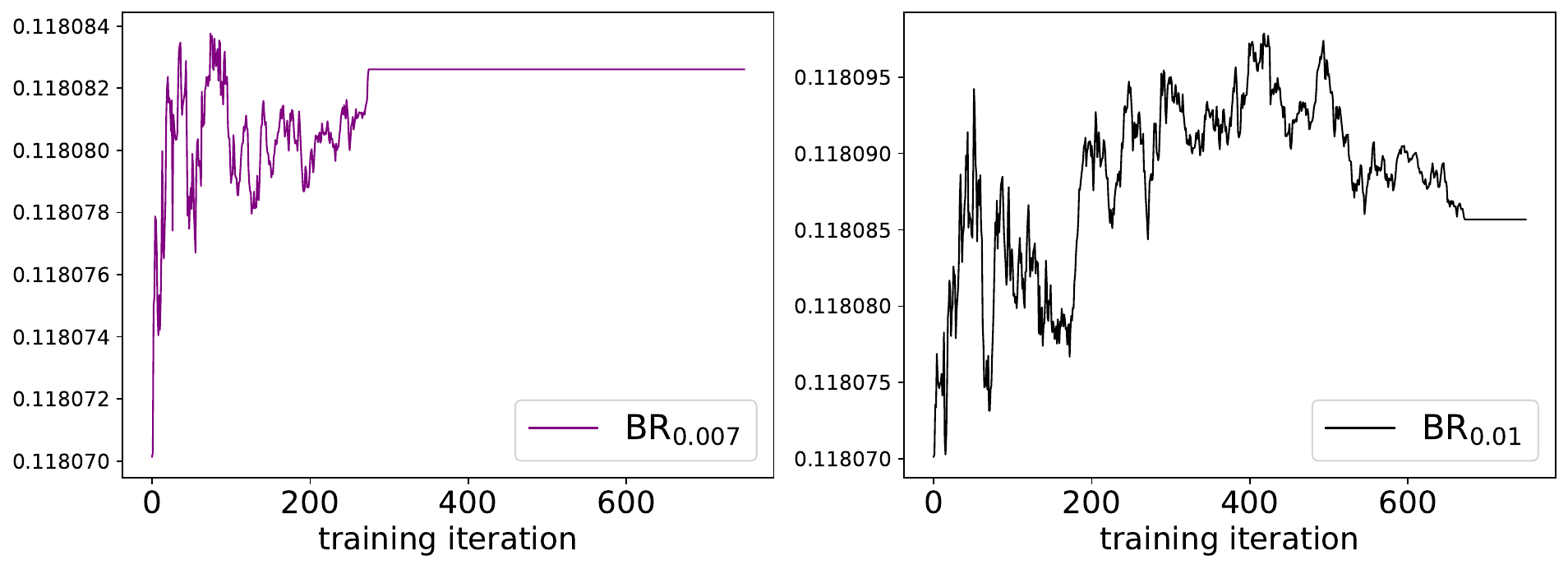}
\endminipage
\caption{Trajectory of the statistical scaling factors from the $10^{\rm th}$ transformer block in a 2-bit DeiT-S throughout the confidence-guided annealing ($\rm CGA$) period with 4 different boundary range settings ($[{\rm BR}_{0.003}$, ${\rm BR}_{0.005}$, ${\rm BR}_{0.007}$, ${\rm BR}_{0.01}]$). The y-axis represents the value of $\alpha_s$. Refer to Appendix. \ref{appendix:complete_alphavis} for the visualization of 3/4-bit DeiT-S.}
\label{fig:cga_dif_br_vis}
\end{center}
\end{figure}

In this section, we first examine the solitary effectiveness of each of the three proposed methods (\textit{i.e.}, ${\rm StatsQ}$, $\rm CGA$, and $\rm QKR$) on DeiT-S quantization. As shown in Table \ref{tab:ablation1}, simply replacing \cite{esser2019learned} with ${\rm StatsQ}$ improves the accuracy by 6.3\%, 0.8\% and 1.0\% in 2/3/4-bit settings, respectively. Moreover, $\rm QKR$ and $\rm CGA$, when applied with ${\rm StatsQ}$, can provide non-negligible improvement across bit-widths. For example, adding $\rm QKR$ improves 0.7\% / 0.59\% and adding $\rm CGA$ improves 0.77\% / 0.44\% in 2-bit / 3-bit settings. All three methods work collaboratively, and the final model with a combination of three methods boosts the accuracy by 7.72\% on 2-bit DeiT-S compared to the baseline $\rm LSQ$.

We then investigate the robustness of boundary range (${\rm BR}_{\rm x}$) selection in $\rm CGA$ on 2-bit quantized DeiT-S with 4 different settings $[{\rm BR}_{0.003}, {\rm BR}_{0.005}, {\rm BR}_{0.007}, {\rm BR}_{0.01}]$. In Table \ref{tab:ablation2}, we observe that ${\rm BR}_{0.005}$ brings the biggest accuracy improvement of 0.72\% while ${\rm BR}_{0.003}$, ${\rm BR}_{0.007}$ and ${\rm BR}_{0.01}$ improve the accuracy by $\sim$ 0.6\%. We further plot the trajectories of the statistical scaling factor with different $\rm BR$ settings, and Fig \ref{fig:cga_dif_br_vis} illustrates that the larger the boundary range is, the more training iterations are required for the oscillating weights to exit the boundary range completely. The result implies that if ${\rm BR}_{\rm x}$ is set too capacious, it will incur longer optimization time and would raise the risk of affecting ``\textit{high confident}'' weights; on the contrary, if ${\rm BR}_{\rm x}$ is too narrow, part of the ``\textit{low confident}'' weights will be frozen, which restricts the effectiveness of $\rm CGA$ on updating oscillating weights. We empirically set the boundary range approximated to the average gradient in the optimized model, and Table \ref{tab:ablation2} shows that $\rm CGA$ is pretty robust to the ${\rm BR}_{\rm x}$ selections as long as ${\rm BR}_{\rm x}$ is not set too aggressively.

\section{Conclusion}
\label{conclusion}
In this work, we use Vision Transformer (ViT) as a study case for investigating quantization oscillation. We uncover the negative influence of learnable scaling factors on escalating weight oscillation. Moreover, we find that the quantized \textit{query} and \textit{key} of self-attention in ViT has a negative mutual influence that intensifies the weight oscillation collectively. In light of our observations, we introduce statistical weight quantization ($\rm StatsQ$), confidence-guided annealing ($\rm CGA$), and \textit{query}-\textit{key} reparameterization ($\rm QKR$) to mitigate the oscillation of quantized weights. With $\rm StatsQ$ and $\rm QKR$, quantization-aware training becomes more stable and converges to a better local minima; with $\rm CGA$, oscillation is further subsided towards the end of the training and ultimately arrives at an oscillation-free model. For future work, we wish to investigate the oscillation phenomenon in more variety of DNNs on diverse applications to examine the effectiveness and generalizability of the proposed methods. 

\section{Acknowledgements}
\label{acknowledgements}
This research was supported by ACCESS - AI Chip Center for Emerging Smart Systems, sponsored by InnoHK funding, Hong Kong SAR, and HKSAR RGC General Research Fund (GRF) No.16203319.

\bibliography{example_paper}

\begin{thebibliography}{33}
\providecommand{\natexlab}[1]{#1}
\providecommand{\url}[1]{\texttt{#1}}
\expandafter\ifx\csname urlstyle\endcsname\relax
  \providecommand{\doi}[1]{doi: #1}\else
  \providecommand{\doi}{doi: \begingroup \urlstyle{rm}\Url}\fi

\bibitem[Banner et~al.(2019)Banner, Nahshan, and Soudry]{banner2019post}
Banner, R., Nahshan, Y., and Soudry, D.
\newblock Post training 4-bit quantization of convolutional networks for
  rapid-deployment.
\newblock \emph{Advances in Neural Information Processing Systems}, 32, 2019.

\bibitem[Bengio et~al.(2013)Bengio, L{\'e}onard, and
  Courville]{bengio2013estimating}
Bengio, Y., L{\'e}onard, N., and Courville, A.
\newblock Estimating or propagating gradients through stochastic neurons for
  conditional computation.
\newblock \emph{arXiv preprint arXiv:1308.3432}, 2013.

\bibitem[Choi et~al.(2018)Choi, Wang, Venkataramani, Chuang, Srinivasan, and
  Gopalakrishnan]{choi2018pact}
Choi, J., Wang, Z., Venkataramani, S., Chuang, P. I.-J., Srinivasan, V., and
  Gopalakrishnan, K.
\newblock Pact: Parameterized clipping activation for quantized neural
  networks.
\newblock \emph{arXiv preprint arXiv:1805.06085}, 2018.

\bibitem[Dosovitskiy et~al.(2021)Dosovitskiy, Beyer, Kolesnikov, Weissenborn,
  Zhai, Unterthiner, Dehghani, Minderer, Heigold, Gelly, Uszkoreit, and
  Houlsby]{dosovitskiy2020image}
Dosovitskiy, A., Beyer, L., Kolesnikov, A., Weissenborn, D., Zhai, X.,
  Unterthiner, T., Dehghani, M., Minderer, M., Heigold, G., Gelly, S.,
  Uszkoreit, J., and Houlsby, N.
\newblock An image is worth 16x16 words: Transformers for image recognition at
  scale.
\newblock In \emph{International Conference on Learning Representations}, 2021.

\bibitem[Esser et~al.(2020)Esser, McKinstry, Bablani, Appuswamy, and
  Modha]{esser2019learned}
Esser, S.~K., McKinstry, J.~L., Bablani, D., Appuswamy, R., and Modha, D.~S.
\newblock Learned step size quantization.
\newblock In \emph{International Conference on Learning Representations}, 2020.

\bibitem[Gong et~al.(2019)Gong, Liu, Jiang, Li, Hu, Lin, Yu, and
  Yan]{gong2019differentiable}
Gong, R., Liu, X., Jiang, S., Li, T., Hu, P., Lin, J., Yu, F., and Yan, J.
\newblock Differentiable soft quantization: Bridging full-precision and low-bit
  neural networks.
\newblock In \emph{Proceedings of the IEEE/CVF International Conference on
  Computer Vision}, pp.\  4852--4861, 2019.

\bibitem[He et~al.(2016)He, Zhang, Ren, and Sun]{he2016deep}
He, K., Zhang, X., Ren, S., and Sun, J.
\newblock Deep residual learning for image recognition.
\newblock In \emph{Proceedings of the IEEE conference on computer vision and
  pattern recognition}, pp.\  770--778, 2016.

\bibitem[Helwegen et~al.(2019)Helwegen, Widdicombe, Geiger, Liu, Cheng, and
  Nusselder]{helwegen2019latent}
Helwegen, K., Widdicombe, J., Geiger, L., Liu, Z., Cheng, K.-T., and Nusselder,
  R.
\newblock Latent weights do not exist: Rethinking binarized neural network
  optimization.
\newblock \emph{Advances in neural information processing systems}, 32, 2019.

\bibitem[Kenton \& Toutanova(2019)Kenton and Toutanova]{devlin2018bert}
Kenton, J. D. M.-W.~C. and Toutanova, L.~K.
\newblock Bert: Pre-training of deep bidirectional transformers for language
  understanding.
\newblock In \emph{Proceedings of naacL-HLT}, pp.\  4171--4186, 2019.

\bibitem[Krizhevsky et~al.(2017)Krizhevsky, Sutskever, and
  Hinton]{imagenet2012}
Krizhevsky, A., Sutskever, I., and Hinton, G.~E.
\newblock Imagenet classification with deep convolutional neural networks.
\newblock \emph{Commun. ACM}, 60\penalty0 (6):\penalty0 84–90, may 2017.
\newblock ISSN 0001-0782.
\newblock \doi{10.1145/3065386}.
\newblock URL \url{https://doi.org/10.1145/3065386}.

\bibitem[Li et~al.(2020)Li, Dong, and Wang]{li2019additive}
Li, Y., Dong, X., and Wang, W.
\newblock Additive powers-of-two quantization: An efficient non-uniform
  discretization for neural networks.
\newblock In \emph{International Conference on Learning Representations}, 2020.

\bibitem[Li et~al.(2022{\natexlab{a}})Li, Xu, Zhang, Cao, Gao, and
  Guo]{liy2022q}
Li, Y., Xu, S., Zhang, B., Cao, X., Gao, P., and Guo, G.
\newblock Q-vit: Accurate and fully quantized low-bit vision transformer.
\newblock In \emph{Advances in Neural Information Processing Systems},
  2022{\natexlab{a}}.

\bibitem[Li \& Gu(2022)Li and Gu]{liz2022vit}
Li, Z. and Gu, Q.
\newblock I-vit: integer-only quantization for efficient vision transformer
  inference.
\newblock \emph{arXiv preprint arXiv:2207.01405}, 2022.

\bibitem[Li et~al.(2022{\natexlab{b}})Li, Yang, Wang, and Cheng]{liz2022qmix}
Li, Z., Yang, T., Wang, P., and Cheng, J.
\newblock Q-vit: Fully differentiable quantization for vision transformer.
\newblock \emph{arXiv preprint arXiv:2201.07703}, 2022{\natexlab{b}}.

\bibitem[Liu et~al.(2018)Liu, Wu, Luo, Yang, Liu, and Cheng]{liu2018bi}
Liu, Z., Wu, B., Luo, W., Yang, X., Liu, W., and Cheng, K.-T.
\newblock Bi-real net: Enhancing the performance of 1-bit cnns with improved
  representational capability and advanced training algorithm.
\newblock In \emph{Proceedings of the European conference on computer vision
  (ECCV)}, pp.\  722--737, 2018.

\bibitem[Liu et~al.(2019)Liu, Mu, Zhang, Guo, Yang, Cheng, and
  Sun]{liu2019metapruning}
Liu, Z., Mu, H., Zhang, X., Guo, Z., Yang, X., Cheng, K.-T., and Sun, J.
\newblock Metapruning: Meta learning for automatic neural network channel
  pruning.
\newblock In \emph{Proceedings of the IEEE/CVF international conference on
  computer vision}, pp.\  3296--3305, 2019.

\bibitem[Liu et~al.(2021{\natexlab{a}})Liu, Lin, Cao, Hu, Wei, Zhang, Lin, and
  Guo]{liu2021swin}
Liu, Z., Lin, Y., Cao, Y., Hu, H., Wei, Y., Zhang, Z., Lin, S., and Guo, B.
\newblock Swin transformer: Hierarchical vision transformer using shifted
  windows.
\newblock In \emph{Proceedings of the IEEE/CVF international conference on
  computer vision}, pp.\  10012--10022, 2021{\natexlab{a}}.

\bibitem[Liu et~al.(2021{\natexlab{b}})Liu, Shen, Li, Helwegen, Huang, and
  Cheng]{liu2021adam}
Liu, Z., Shen, Z., Li, S., Helwegen, K., Huang, D., and Cheng, K.-T.
\newblock How do adam and training strategies help bnns optimization.
\newblock In \emph{International Conference on Machine Learning}, pp.\
  6936--6946. PMLR, 2021{\natexlab{b}}.

\bibitem[Liu et~al.(2021{\natexlab{c}})Liu, Wang, Han, Zhang, Ma, and
  Gao]{liu2021post}
Liu, Z., Wang, Y., Han, K., Zhang, W., Ma, S., and Gao, W.
\newblock Post-training quantization for vision transformer.
\newblock \emph{Advances in Neural Information Processing Systems},
  34:\penalty0 28092--28103, 2021{\natexlab{c}}.

\bibitem[Liu et~al.(2022)Liu, Cheng, Huang, Xing, and Shen]{liu2022nonuniform}
Liu, Z., Cheng, K.-T., Huang, D., Xing, E.~P., and Shen, Z.
\newblock Nonuniform-to-uniform quantization: Towards accurate quantization via
  generalized straight-through estimation.
\newblock In \emph{Proceedings of the IEEE/CVF Conference on Computer Vision
  and Pattern Recognition}, pp.\  4942--4952, 2022.

\bibitem[Miyashita et~al.(2016)Miyashita, Lee, and Murmann]{miyashita2016pot}
Miyashita, D., Lee, E.~H., and Murmann, B.
\newblock Convolutional neural networks using logarithmic data representation.
\newblock \emph{arXiv preprint arXiv:1603.01025}, 2016.

\bibitem[Nagel et~al.(2019)Nagel, Baalen, Blankevoort, and
  Welling]{nagel2019data}
Nagel, M., Baalen, M.~v., Blankevoort, T., and Welling, M.
\newblock Data-free quantization through weight equalization and bias
  correction.
\newblock In \emph{Proceedings of the IEEE/CVF International Conference on
  Computer Vision}, pp.\  1325--1334, 2019.

\bibitem[Nagel et~al.(2020)Nagel, Amjad, Van~Baalen, Louizos, and
  Blankevoort]{nagel2020up}
Nagel, M., Amjad, R.~A., Van~Baalen, M., Louizos, C., and Blankevoort, T.
\newblock Up or down? adaptive rounding for post-training quantization.
\newblock In \emph{International Conference on Machine Learning}, pp.\
  7197--7206. PMLR, 2020.

\bibitem[Nagel et~al.(2022)Nagel, Fournarakis, Bondarenko, and
  Blankevoort]{nagel2022overcoming}
Nagel, M., Fournarakis, M., Bondarenko, Y., and Blankevoort, T.
\newblock Overcoming oscillations in quantization-aware training.
\newblock In Chaudhuri, K., Jegelka, S., Song, L., Szepesvari, C., Niu, G., and
  Sabato, S. (eds.), \emph{Proceedings of the 39th International Conference on
  Machine Learning}, volume 162 of \emph{Proceedings of Machine Learning
  Research}, pp.\  16318--16330. PMLR, 17--23 Jul 2022.

\bibitem[Touvron et~al.(2021)Touvron, Cord, Douze, Massa, Sablayrolles, and
  J{\'e}gou]{touvron2021training}
Touvron, H., Cord, M., Douze, M., Massa, F., Sablayrolles, A., and J{\'e}gou,
  H.
\newblock Training data-efficient image transformers \& distillation through
  attention.
\newblock In \emph{International Conference on Machine Learning}, pp.\
  10347--10357. PMLR, 2021.

\bibitem[Vaswani et~al.(2017)Vaswani, Shazeer, Parmar, Uszkoreit, Jones, Gomez,
  Kaiser, and Polosukhin]{vaswani2017attention}
Vaswani, A., Shazeer, N., Parmar, N., Uszkoreit, J., Jones, L., Gomez, A.~N.,
  Kaiser, {\L}., and Polosukhin, I.
\newblock Attention is all you need.
\newblock \emph{Advances in neural information processing systems}, 30, 2017.

\bibitem[Wu et~al.(2019)Wu, Dai, Zhang, Wang, Sun, Wu, Tian, Vajda, Jia, and
  Keutzer]{wu2019fbnet}
Wu, B., Dai, X., Zhang, P., Wang, Y., Sun, F., Wu, Y., Tian, Y., Vajda, P.,
  Jia, Y., and Keutzer, K.
\newblock Fbnet: Hardware-aware efficient convnet design via differentiable
  neural architecture search.
\newblock In \emph{Proceedings of the IEEE/CVF Conference on Computer Vision
  and Pattern Recognition}, pp.\  10734--10742, 2019.

\bibitem[Xiao et~al.(2022)Xiao, Lin, Seznec, Wu, Demouth, and
  Han]{xiao2022smoothquant}
Xiao, G., Lin, J., Seznec, M., Wu, H., Demouth, J., and Han, S.
\newblock Smoothquant: Accurate and efficient post-training quantization for
  large language models.
\newblock \emph{arXiv}, 2022.

\bibitem[Yun et~al.(2019)Yun, Han, Oh, Chun, Choe, and Yoo]{yun2019cutmix}
Yun, S., Han, D., Oh, S.~J., Chun, S., Choe, J., and Yoo, Y.
\newblock Cutmix: Regularization strategy to train strong classifiers with
  localizable features.
\newblock In \emph{Proceedings of the IEEE/CVF international conference on
  computer vision}, pp.\  6023--6032, 2019.

\bibitem[Zhang et~al.(2018{\natexlab{a}})Zhang, Yang, Ye, and Hua]{zhang2018lq}
Zhang, D., Yang, J., Ye, D., and Hua, G.
\newblock Lq-nets: Learned quantization for highly accurate and compact deep
  neural networks.
\newblock In \emph{Proceedings of the European conference on computer vision
  (ECCV)}, pp.\  365--382, 2018{\natexlab{a}}.

\bibitem[Zhang et~al.(2018{\natexlab{b}})Zhang, Cisse, Dauphin, and
  Lopez-Paz]{zhang2017mixup}
Zhang, H., Cisse, M., Dauphin, Y.~N., and Lopez-Paz, D.
\newblock mixup: Beyond empirical risk minimization.
\newblock In \emph{International Conference on Learning Representations},
  2018{\natexlab{b}}.

\bibitem[Zhou et~al.(2016)Zhou, Wu, Ni, Zhou, Wen, and Zou]{zhou2016dorefa}
Zhou, S., Wu, Y., Ni, Z., Zhou, X., Wen, H., and Zou, Y.
\newblock Dorefa-net: Training low bitwidth convolutional neural networks with
  low bitwidth gradients.
\newblock \emph{arXiv preprint arXiv:1606.06160}, 2016.

\bibitem[Zhu et~al.(2020)Zhu, Duong, and Liu]{zhu2020xor}
Zhu, S., Duong, L.~H., and Liu, W.
\newblock Xor-net: an efficient computation pipeline for binary neural network
  inference on edge devices.
\newblock In \emph{2020 IEEE 26th International Conference on Parallel and
  Distributed Systems (ICPADS)}, pp.\  124--131. IEEE, 2020.

\end{thebibliography}
\bibliographystyle{icml2023}

\onecolumn
\clearpage
\section{Appendix}
\subsection{Progression of $\alpha_s$ in $\rm CGA$ with Different ${\rm BR}_{\rm x}$}
\label{appendix:complete_alphavis}
\begin{figure*}[h!]
\centering
\includegraphics[width=17cm]{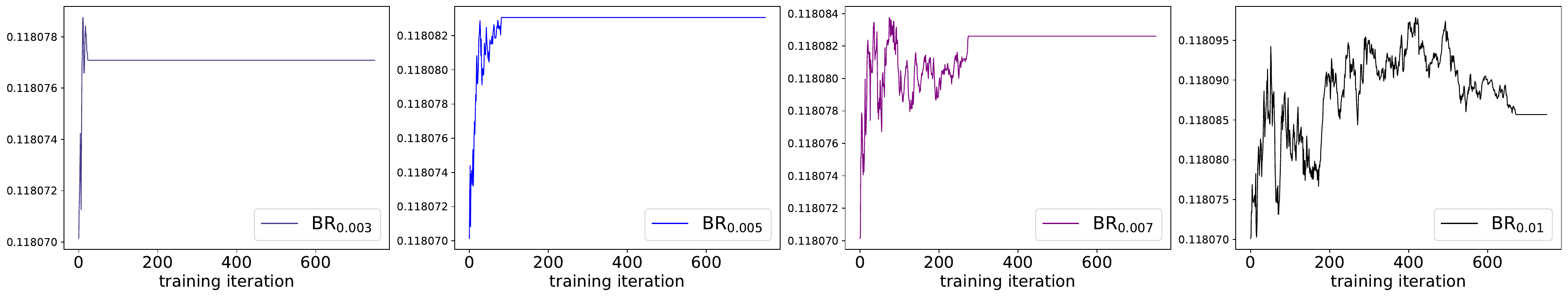}
\caption{Trajectory of statistical scaling factors $\alpha_s$ from the $10^{\rm th}$ transformer blocks in a \textbf{2-bit} DeiT-S throughout $\rm CGA$ with 4 different boundary ranges $[{\rm BR}_{0.003}, {\rm BR}_{0.005}, {\rm BR}_{0.007}, {\rm BR}_{0.01}]$. The y-axis represents the value of $\alpha_s$.}
\label{fig:alpha_s_2bit}
\end{figure*}

\begin{figure*}[h!]
\centering
\includegraphics[width=17cm]{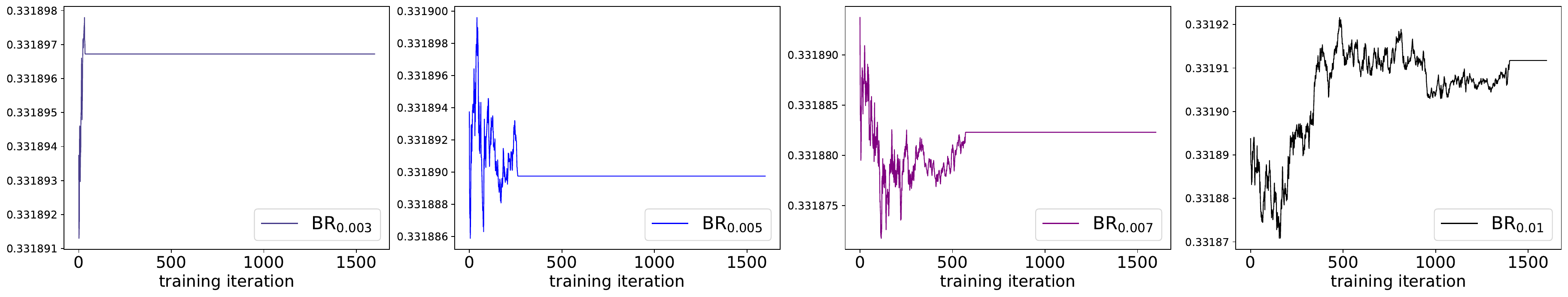}
\caption{Trajectory of statistical scaling factors $\alpha_s$ from the $10^{\rm th}$ transformer blocks in a \textbf{3-bit} DeiT-S throughout $\rm CGA$ with 4 different boundary ranges $[{\rm BR}_{0.003}, {\rm BR}_{0.005}, {\rm BR}_{0.007}, {\rm BR}_{0.01}]$. The y-axis represents the value of $\alpha_s$.}
\label{fig:alpha_s_3bit}
\end{figure*}

\begin{figure*}[h!]
\centering
\includegraphics[width=17cm]{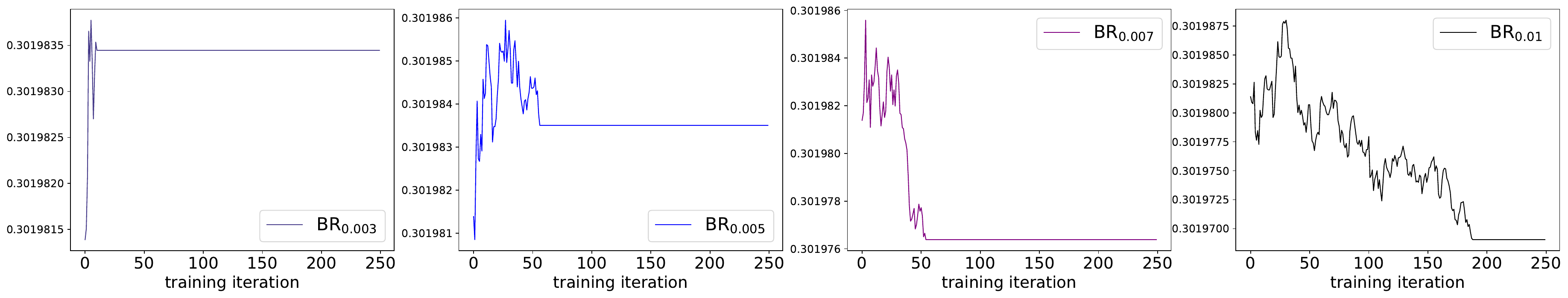}
\caption{Trajectory of statistical scaling factors $\alpha_s$ from the $10^{\rm th}$ transformer blocks in a \textbf{4-bit} DeiT-S throughout $\rm CGA$ with 4 different boundary ranges $[{\rm BR}_{0.003}, {\rm BR}_{0.005}, {\rm BR}_{0.007}, {\rm BR}_{0.01}]$. The y-axis represents the value of $\alpha_s$.}
\label{fig:alpha_s_4bit}
\end{figure*}

In this section, we provide the trajectory visualization of statistical scaling factors $\alpha_s$ from the $10^{\rm th}$ transformer blocks throughout confidence-guided annealing ($\rm CGA$) period with different boundary ranges [${\rm BR}_{0.003}$, ${\rm BR}_{0.005}$, ${\rm BR}_{0.007}$, ${\rm BR}_{0.01}$] and compare their different behaviors in 2/3/4-bit DeiT-S. From Fig.~\ref{fig:alpha_s_2bit}-\ref{fig:alpha_s_4bit}, $\rm CGA$ demonstrates its capability of successfully guiding models to become oscillation-free regardless of model bit-widths and ${\rm BR}_{x}$. Moreover, it is evident that more training iterations are required for all the weights to exit a larger boundary range across different bit-widths. Additionally, we can observe that fewer training iterations are needed for all the weights to stop oscillating in a 4-bit model than 2-bit/3-bit models, \textit{e.g.}, $\sim$250 iterations for 4-bit DeiT-S, $\sim$1500 iterations for 3-bit DeiT-S and $\sim$700 iterations for 2-bit DeiT-S. This observation aligns with our findings that oscillation is less detrimental in the 4-bit setting due to the higher resolution of the model.

\end{document}